%% file: paper.tex
\documentclass[runningheads]{llncs}
\usepackage[hidelinks]{hyperref}
\usepackage[T1]{fontenc}
\usepackage{graphicx}
\usepackage{booktabs}
\usepackage[misc]{ifsym}
\newcommand{\corr}{(\Letter)}

\usepackage{amsmath}
\usepackage{amsfonts}
\usepackage[x11names]{xcolor}
\usepackage{appendix}
\usepackage[most]{tcolorbox}
\usepackage{tcolorbox}
\usepackage{tabularx}
\tcbuselibrary{listings,breakable}
\usepackage[detect-all,
            round-mode=figures,
            round-precision=3,
            scientific-notation = engineering]{siunitx}
\usepackage{array}
\usepackage[section]{placeins}
\usepackage[compatibility=false]{caption}
\usepackage{subcaption}
\usepackage{float}
\usepackage{graphicx}
\usepackage{svg}
\usepackage{orcidlink}

\begin{document}

\title{Standard vs. Modular Sampling: Best Practices for Reliable LLM Unlearning}

\author{Praveen Bushipaka \inst{1,2} \corr \orcidlink{0009-0009-7753-8662} \and
Lucia Passaro \inst{2}  \orcidlink{0000-0003-4934-5344} \and
Tommaso Cucinotta \inst{1} \orcidlink{0000-0002-0362-0657}}

\authorrunning{P.Bushipaka et al.}
\institute{Scuola Superiore Sant'Anna  \email{\{name.surname\}@santannapisa.it},
\and
University of Pisa \email{lucia.passaro@unipi.it},
\email{praveen.bushipaka@phd.unipi.it}
}

\maketitle     
\begin{center}
\small
Preprint of the chapter published in
\emph{Machine Learning and Principles and Practice of Knowledge Discovery in Databases (ECML PKDD 2025 Workshops)},
Communications in Computer and Information Science (CCIS), vol. 2840, Springer, 2026.\\
DOI: \url{https://doi.org/10.1007/978-3-032-19099-4_43}
\end{center}
\begin{abstract}
A conventional LLM Unlearning setting consists of two subsets -"forget" and "retain", with the objectives of removing the undesired knowledge from the forget set while preserving the remaining knowledge from the retain. In privacy-focused unlearning research, a retain set is often further divided into neighbor sets, containing either directly or indirectly connected to the forget targets; and augmented by a general-knowledge set. A common practice in existing benchmarks is to employ only a single neighbor set, with general knowledge which fails to reflect the real-world data complexities and relationships. LLM Unlearning typically involves 1:1 sampling or cyclic iteration sampling. However, the efficacy and stability of these de facto standards have not been critically examined. In this study, we systematically evaluate these common practices. Our findings reveal that relying on a single neighbor set is suboptimal and that a standard sampling approach can obscure performance trade-offs. Based on this analysis, we propose and validate an initial set of best practices: \textbf{(1)} Incorporation of diverse neighbor sets to balance forget efficacy and model utility, \textbf{(2)} Standard 1:1 sampling methods are inefficient and yield poor results, \textbf{(3)} Our proposed \textbf{Modular Entity-Level Unlearning (MELU)} strategy as an alternative to cyclic sampling. We demonstrate that this modular approach, combined with robust algorithms, provides a clear and stable path towards effective unlearning. Our code can be found at \hyperlink{MELU}{\url{https://github.com/praveensonu/MELU}}. 
\end{abstract}

\input{ECML_PKDD_2025_Author_Kit/introduction}

\input{ECML_PKDD_2025_Author_Kit/related_works}

\input{ECML_PKDD_2025_Author_Kit/experimental_setup}

\input{ECML_PKDD_2025_Author_Kit/results_discussion}

\input{ECML_PKDD_2025_Author_Kit/conclusion}


%
%
%
\FloatBarrier
\bibliographystyle{splncs04}
\bibliography{references}





\appendix
\section{Appendix}
\input{ECML_PKDD_2025_Author_Kit/appendix}

\end{document}

%% file: ECML_PKDD_2025_Author_Kit/introduction.tex
\section{Introduction}


\subsubsection{Large Language Models (LLMs)} \cite{10.5555/3295222.3295349,10.5555/3495724.3495883} are trained on vast amounts of data scraped from the web, enabling them to process billions of learnable parameters. This extensive scaling enables them to address a wide array of complex linguistic tasks, exhibiting performance that approaches human-level proficiency in both language understanding and generation. However, this scale introduces a significant challenge: the models can memorize sensitive information such as personal data, copyrighted content, harmful data and output this information \cite{274574,9157084}, raising concerns over their potential misuse \cite{staab2024beyond}. 


\paragraph{}To address this problem, \textbf{LLM Unlearning} \cite{yao2024large,miranda2025preserving} has emerged as a promising technique, aiming to remove specific knowledge and abilities while preserving the overall integrity and performance of the model. The conventional approach to LLM Unlearning involves two primary goals: (1) the unlearning process should remove the specified target knowledge and its associated abilities; (2) the unlearning must respect the model integrity and must not affect the non-target model abilities even if they are directly or indirectly related to the target \cite{liu2025rethinking}. For instance, if the target knowledge includes information about an author \textit{Benedetto Varchi} who was born in \textit{Florence, Italy} then the unlearning process should successfully remove the Benedetto Varchi's association to Florence, while retaining all the other knowledge about Florence (such as its connection of a city in Italy). To achieve these objectives, unlearning usually involves two datasets: a \textbf{Forget set}, containing the knowledge that needs to be erased and a \textbf{Retain set}, containing the knowledge that needs to be preserved. The process generally maximizes loss on the forget set and minimizes it on the retain set, helping to avoid issues like \textbf{Degeneration Behavior} and \textbf{Catastrophic Forgetting}.

\paragraph{} As LLM Unlearning transitions from a theoretical concept to a practical tool, best practices for adopting unlearning need to be looked into. Researchers often construct a retain set using a 1:1 ratio of forget-to-retain samples \cite{maini2024tofu,premptis-etal-2025-ails}, drawing from a single type of neighbor set and a general knowledge pool \cite{maini2024tofu}. When the retain set is larger, a cyclic sampling approach is employed to pair the forget and retain data during the unlearning process \cite{premptis-etal-2025-ails,jang-etal-2023-knowledge}. While these practices offer a straightforward path, their impact on the goal of unlearning is poorly understood. Do these simple heuristics represent an optimal strategy, or do they introduce hidden risks and performance ceilings?


\paragraph{}In this work, we make a step towards establishing a set of evidence-based best practices for LLM Unlearning. First, we look into the dataset creation practice, by extending the Wikipedia Person Unlearning (WPU) \cite{liu-etal-2024-revisiting} dataset by incorporating multiple neighbor sets into it. We analyze common data configurations (such as only direct or indirect neighbors) and sampling methods (1:1 sampling, cyclic), identify their strengths and weaknesses, and propose our strategy as an alternative.

\paragraph{}More precisely, \textbf{our contributions} are as follows:
\begin{enumerate}
    \item  \textbf{A Critical Analysis of Common Data Practices}: A systematic evaluation of how retain set composition impacts unlearning outcomes. We demonstrate how a diverse retain set is crucial for balancing Forget Efficacy and Model Utility. 
    \item \textbf{Comparison of common sampling strategies of LLM Unlearning}: A comparative analysis of common sampling methods for LLM Unlearning. We discover that the common practice of 1:1 sampling is ineffective.   
    \item \textbf{Proposal of MELU as a sampling technique}: We introduce \textbf{Modular Entity-Level Unlearning (MELU)}, a simple structured sampling strategy, that demonstrates more stable and effective Unlearning than conventional cyclic sampling. 
\end{enumerate}

%% file: ECML_PKDD_2025_Author_Kit/related_works.tex
\section{Preliminaries}
\subsection{Unlearning in Large Language Models}

Prior works of unlearning in Large Language Models focus on classification tasks~\cite{jang-etal-2023-knowledge}, but due to the increase in adoption of generative AI in industries and everyday life, especially chatbots and instruction-tuned LLMs, the research focus has moved to question-and-answer (Q\&A) tasks. 

Given the Large Language model \textit{M} with its parameters $\theta$, the forget set  $\mathcal{D}_f = \{x_f, y_f\}$ contains the samples that need to be forgotten by $\mathcal{\textit{M}(;\theta)}$ and the retain set $\mathcal{D}_r = \{x_r, y_r\}$ those $\mathcal{\textit{M}(;\theta)}$ needs to preserve, where ${x, y}$ are questions and answers (inputs and their labels) in the LLM Unlearning task. Our goal is to provide an updated/forgotten LLM with parameters $\theta_*$ satisfying the objectives mentioned previously. 

Although designs vary, most fine-tuning based unlearning algorithms objectives can be mathematically written as \cite{ji2024reversing}

\[
\underset{\theta_*}{\min} L(\theta_*) = \underset{\theta_*}{\min} \left( -L_f(\theta_*) + \lambda L_r(\theta_*) \right)
\]

The equation provides the objectives, the first loss term - forget loss $\mathcal{L}_f(\theta_*)$ maximizes the loss on forget set, and the second loss term - retain loss $\mathcal{L}_R(\theta_*)$ minimizes the loss on the retain set. $\lambda$ is a hyper-parameter controlling the retain strength.

\subsubsection{Entity Unlearning}
There are two types of unlearning: \textit{Instance-Level Unlearning} and \textit{Entity-Level Unlearning} \cite{choi-etal-2025-opt,maini2024tofu,liu-etal-2024-revisiting}: the former erases specific knowledge about a forget-target, whereas the latter removes all knowledge of that entity (e.g. a person, institution, book series etc). Formally, given entities $\epsilon = {e_1,e_2..e_n}$ to forget, each $e_i$ is represented by Q\&A pairs $\textit{e}_i = \{(x_{i1}, y_{i1}) ... (x_{in}, y_{in})\}$. The model $M(\theta)$, is trained on dataset $\textit{D}$, is split into forget set $\textit{D}_f$ and a disjoint retain set $\textit{D}_r = \textit{D} \, \backslash\, \textit{D}_f$. In this work, we focus on Entity-level unlearning.

\subsection{Datasets}

Since $\textit{D}_r$ is disjoint from $\textit{D}_f$, it could contain potentially all the pretrained data excluding the $\textit{D}_f$. This is impractical to implement, and prior works address this challenge by assessing performance on general knowledge benchmarks such as MMLU \cite{wang2024mmluprorobustchallengingmultitask} or creating an entirely new general knowledge dataset \cite{pmlr-v235-li24bc,maini2024tofu,liu-etal-2024-revisiting} and creating neighbor sets, which are subsets of $\textit{D}_r$ expected to be influenced by the unlearning process. These neighbor sets are constructed based on the assumption that data points similar to $\textit{D}_f$ or involved in unlearning are more likely to be impacted during unlearning. From the literature, we identify three types of neighbor sets:

\subsubsection{Direct Neighbor set $(\textit{N}_{d})$ -} Direct Neighbor sets contain the entities that are closely associated and directly connected to $\textit{D}_f$ \cite{liu-etal-2024-revisiting,jin2024rwku}. These include but are not limited to place of birth, family tree, education, personal achievements, and everything that is directly linked to the forget target. For instance, for a forget target - '\textit{Benedetto Varchi} was born in \textit{Florence}', and information on Florence is considered as a part of its corresponding Direct Neighbor set, assuming this is directly influenced due to the forgetting of Benedetto Varchi's birthplace knowledge.

\subsubsection{Indirect Neighbor set $(\textit{N}_{ind})$ -} First introduced in TOFU \cite{maini2024tofu}, an indirect set consists of entities sharing a semantic or contextual relationship with the forget target, without being directly linked. These connections may be based on historical period, domain, ideology, or thematic relevance-not necessarily profession. For example, if the entity is \textit{Benedetto Varchi's}, an Italian humanist and historian of the fifteenth century, the corresponding Indirect neighbor set consists of data on Leonardo Bruni, Francesco Petrarca etc. who were also Italian historians of the similar period. It is difficult to derive the indirect connections without looking at the model activations and pre-trained dataset. So, for our study, we use the already proposed approach of profession to be the indirect connection.

\subsubsection{Syntactic similarity $(\textit{N}_s)$ -} Introduced by \cite{chang-lee-2025-retain}, they expand the present neighbor sets to syntactic similarity neighbor set, showing that syntactic similarity is the most influenced due to the nature of question-answering unlearning task. For example, 'When was Benedetto Varchi born?' can have influence on a question with similar syntax such as 'When was Donald Trump born?'. To avoid this, they propose an entirely new neighbor set.

\subsection{LLM Unlearning Practices}

We do not discuss LLM Unlearning algorithms, rather we discuss their implementations. For unlearning algorithms please refer Appendix:\ref{appendix:methods}.

\subsubsection{Batch and Sequential Unlearning}

\paragraph{} \textbf{Batch Unlearning}, commonly used refers to unlearning the model on all the forget targets at once. While straightforward, this approach has been observed to suffer from instability and leads to catastrophic collapse \cite{jang-etal-2023-knowledge}. \textbf{Sequential Unlearning}, proposed by \cite{jang-etal-2023-knowledge} and further extended by  \cite{premptis-etal-2025-ails}, divides the forget set into chunks, processing each chunk independently and simultaneously processing the retain data, making it ideal for a realistic setting. 

\subsubsection{1:1 and Cyclic Sampling} \label{practices:impl}

\paragraph{} Given the simultaneous maximization of loss on forget sample and minimization on retain sample, a sampling method usually contains how these samples are arranged and how many samples are used in an epoch for the unlearning algorithm. A common sampling practice is \textbf{1:1 Sampling}, i.e., in an epoch, the number of retain samples is not higher than the number of forget samples. There are two ways to do this: \textbf{(Method - a)} creating the dataset with forget and retain samples of the same length -- recent datasets such as the SemEVAL Task-4 competition\footnote{https://llmunlearningsemeval2025.github.io/} follow this structure; \textbf{(Method - b)} randomly choosing the same number of forget and retain samples for every epoch -- initially implemented by \cite{maini2024tofu} and followed by \cite{liu-etal-2024-revisiting,qiu2025datainterconnectivityshapesllms,yuan2025a,fan2025simplicity,mekala-etal-2025-alternate} and many more by reproducing their code, this practice has become common for baselines in LLM Unlearning.

Another common sampling practice is \textbf{Cyclic Sampling} \cite{premptis-etal-2025-ails,jang-etal-2023-knowledge}, in which all the retain samples are utilized by cycling forget samples. As in figure \ref{melu}, a cyclic setting might have a retain sample unrelated to the forget sample. A drawback of this approach is the loss calculation of forget sample with unrelated retain sample. In this study, we introduce \textbf{Modular Entity-Level Unlearning (MELU)} strategy, in which during the unlearning process, each forget target is paired only with its respective retain samples. 

\section{Related Work}

\paragraph{} Current unlearning datasets include either direct $(\textit{N}_d)$ or indirect neighbor($\textit{N}_{ind}$) sets, but never both. \textbf{TOFU} \cite{maini2024tofu} uses 200 synthetic authors as indirect neighbors $(\textit{N}_{ind})$, plus 100 real authors and 117 facts, but omits interconnectivity \cite{qiu2025datainterconnectivityshapesllms} and ignores direct neighbors($(\textit{N}_d)$). \textbf{RWKU} \cite{jin2024rwku} and \textbf{WPU} \cite{liu-etal-2024-revisiting} include only a Direct Neighbor set plus a general knowledge set. \textbf{RWKU}'s focus on 200 high profile figures makes unlearning impractical because their large online footprints means pre-trained LLMs almost certainly have absorbed vast amounts of their data, making it difficult to unlearn; realistic unlearning requests involve individuals with moderate online presence. Datasets from \cite{chang-lee-2025-retain,choi-etal-2025-opt} attempt to include both $(\textit{N}_d)$ and $(\textit{N}_{ind})$ but they rely on bi-directional relationships for $(\textit{N}_d)$, requiring mutual links in their respective Wiki pages, creating a blind spot: e.g., \textit{"Varchi was born in Florence"}, Varchi's page links to Florence, but Florence's page does not link back to Varchi. Although Florence is directly connected and would influence the unlearning process, the bi-directional approach would exclude this from the retain set. Even though these benchmarks exist, LLM unlearning still lacks a standard protocol or methodology for building forget and retain sets, even as unlearning requests become common across applications. \cite{thaker2025positionllmunlearningbenchmarks} provides a new direction on backdrops of the unlearning benchmark datasets. Post-Unlearning, they combine forget and retain queries and ask the model, only to find the model either recognizes them both as forget samples \textit{(outputs IDK)} or retain samples \textit{(outputs correct answers)}.  

In the \textit{1:1 Sampling} \textbf{Method-a} limits the retain set during the data construction, and \textbf{Method-b} limits the retain set during unlearning, especially experiments conducted on \cite{maini2024tofu} benchmark use various splits, fail to leverage the full retain set. For example, TOFU benchmark (4000 samples) has three splits of various forget set sizes (40, 200, 400), leading various retain set sizes (3.96k, 3.8k, 3.6k). TOFU authors unlearn for 5 epochs on these splits. For the largest split of 400 forget samples, at the maximum can attend only 2000 retain samples (if retain samples are sequentially chosen). Albeit, SemEVAL dataset uses \textbf{Method-a}, the winner of the competition \cite{premptis-etal-2025-ails} doesn't follow this approach, instead they follow 1:n-forget:retain approach (in a cyclic sequential unlearning process), making for each batch only for 1 forget sample and n retain samples are present. 

In this paper, we extend the WPU \cite{liu-etal-2024-revisiting} neighbor by adding indirect neighbors with syntactically similar Q\&A's, and a dedicated test set. We further look into the sampling strategies provided by \cite{maini2024tofu}, 1:1 sampling, cyclic sampling in a batch unlearning scenario. We also introduce \textbf{Modular Entity-Level Unlearning (MELU)} strategy, in which each forget target is paired only with its respective retain samples. Our work neither aims to present a benchmark dataset nor an unlearning method, rather improve current best practices in creating LLM Unlearning Benchmark datasets and sampling.

\begin{figure}[t]
\includegraphics[width=\textwidth]{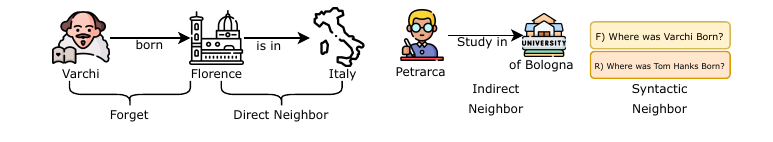}
\caption{Types of Neighbor sets and their connections to the forget sample} \label{neighbor_set}
\end{figure}


%% file: ECML_PKDD_2025_Author_Kit/experimental_setup.tex
\section{Experimental Setup}

\subsection{Dataset Construction}

In this work, we chose the Wikipedia Person Unlearning (WPU)~\cite{liu-etal-2024-revisiting} dataset, which consists of 100 forget targets, Direct neighbor set and general retain set. The dataset is divided into forget\_2, forget\_20 and forget\_100 parts, splitting the \# of target entities to 2, 20 and 100. We chose forget\_20 for the extension of Neighbor sets to $(\textit{N}_{ind})$, $(\textit{N}_s)$ and addition of a test set. The choice of WPU dataset comes from their approach in dataset construction. It was constructed by selecting the least popular Wikipedia people based on their views, these are the forget entities. They construct the $D_r$ through incorporating $(\textit{N}_{d})$ by scraping the hyperlinks that are connected to the Wikipedia page of the person $(\textit{N}_{d})$ and General knowledge set by scraping the Wiki pages of popular people on Wikipedia (General Knowledge set). Other datasets use well known figures information, which might be difficult to unlearn, or a bi-directional approach for $(\textit{N}_{d})$, or synthetically created datasets. WPU stands as an ideal choice for our experiments, as the entities are not well known and has limited online presence providing a realistic unlearning situation.

\subsubsection{Indirect Neighbor set $(\textit{N}_{ind})$ creation -}

For Indirect connection we follow \cite{maini2024tofu}, and chose to find entities of similar profession. Finding similar profession entities for lesser known people was challenging as it requires scraping Google search suggestions, which were often unavailable or inconsistent. To overcome this, we used an LLM to generate similar profession names. Specifically, we chose LLaMA 3.3 70B model \cite{grattafiori2024llama3herdmodels} for this task, since our unlearning experiments were conducted on the LLaMA 3.1 8B Instruct model \cite{grattafiori2024llama3herdmodels}. We assumed that the models of the same family would likely share similar pre-training knowledge. We prompted (Appendix:\ref{appendix:idrect_generation}) the model to generate six names for each forget target, ending up with 120 indirect connections. 

Once we had the Indirect connection entities, we scraped their Wikipedia data and used LLaMA 3.3 70B model to generate the Q\&A's from it. We aimed for at least two questions per section, and instructed the LLM to follow an Interrogative syntactic structure. In total, we ended up with 1409 Q\&A pairs. Then, for each forget target, we randomly picked five indirect connection entities to build the $(\textit{N}_{ind})$- giving us a total of 1144 Q\&A pairs (Appendix:\ref{appendix:indirect}).

\subsubsection{Test set $(\textit{D}_t)$ creation -}

We construct a test with mix of multiple neighbors for evaluation. We use the remaining samples of $(\textit{N}_{ind})$ for indirect connections  and 200 random samples from the general knowledge set for the test set. To create the samples for $(\textit{N}_d)$, we prompted LLaMA 3.3 70B to provide three new basic Q\&A for every answer from forget set. If we are forgetting the link "Adrienne Monnier -> Paris", we made three Q\&A's about Paris (Appendix:\ref{appendix:test_set}). Finally, we created a test set with 738 Q\&A pairs.

Due to the pre-structure of the WPU dataset \cite{liu-etal-2024-revisiting}, which includes $(\textit{N}_d)$ and general knowledge set, we were unable to create a standalone $(\textit{N}_s)$ dataset. Instead we generated the neighbor sets in a similar syntactic manner. To do this, in the Q\&A generation prompt \ref{appendix:qa_generation}, we provided the model to follow interrogative syntactic structure. We verified the syntactic similarity between the forget and retain Q\&A pairs with edit distance algorithm \cite{8054419}, achieved a mean of 40\% similarity.

\begin{figure}[t]
\includegraphics[width=\textwidth]{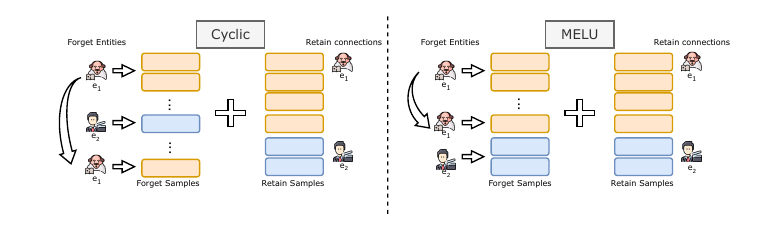}
\caption{MELU setting. In cyclic, each entities ($e_1, e_2.. e_n$) forget samples are cycled on to unrelated ($e_3 \rightarrow e_1$) connections, In MELU entities ($e_1, e_2..e_n$) are cycled only onto their respective target connections ($e_1 \rightarrow e_1$).} \label{melu}
\end{figure}

\subsection{Unlearning Methods}
Before unlearning, we \textbf{fine-tuned} LLaMA 3.1 8B Instruct \cite{grattafiori2024llama3herdmodels} model on all the datasets. We experiment with three unlearning algorithms- Gradient Difference \cite{liu2022continuallearningprivateunlearning} and Negative Preference Optimization \cite{zhang2024negative} for Un-targeted unlearning and Direct Preference Optimization \cite{10.5555/3666122.3668460} for Targeted Unlearning. An Un-targeted unlearning does not contain a replacement sample for the forgetting sample, whereas targeted unlearning does and typically has phrases such as "I don't know".

We compare seven different settings. The initial three are on common data practices, and next four are common implementation practices. We have pool of $|D_f|$ = 98, and $|D_r|$ = 1801 samples. Given that all these experiments revolve around manipulating retain set, we select retain subsets $D_r$ according to:

\textbf{Data Practices\footnote{Given 1:1 sampling (both Sequence and Random) will not utilize the complete $D_r$ during Unlearning, all the Data Practices experiments are conducted in Cyclic Implementation.}}
\begin{enumerate}
    \item \textbf{Direct-Neighbor} $(\textit{N}_d)$. Restrict $D_r$ to only samples with the $(\textit{N}_d)$.
    \item \textbf{Indirect-Neighbor} $(\textit{N}_{ind})$. Restrict $D_r$ to only samples with the $(\textit{N}_{ind})$.
    \item \textbf{Balanced}. We combine $(\textit{N}_d)$ and $(\textit{N}_{ind})$ for our $D_r$. Since our $(\textit{N}_{ind})$ is over sampled, we balance the dataset by selecting an equal number of samples as the $(\textit{N}_d)$ samples for each entity. 
\end{enumerate}

\textbf{Sampling Practices}
\begin{enumerate}
    \item \textbf{1:1 seq}. i.e., We draw $|D_r| = |D_f|$ samples by selecting the top 98 $D_r$ items, matching one retain for each forget. This is the method \textbf{a)} from section \ref{practices:impl}.
    \item \textbf{1:1 random}. i.e., We draw $|D_r| = |D_f|$ samples by randomly selecting 98 $D_r$ items for each epoch, matching one retain for each forget. This is the method \textbf{b)} from section \ref{practices:impl}.
    \item \textbf{Cyclic}. Rotate through the full $D_r$ pool sequentially until we collect 1801 samples - i.e. cycle the 98 sized $|D_f|$ across all the 1801 samples.
    \item \textbf{Modular Entity-Level Unlearning (MELU)}. For each entity or forget target appearing the $D_f$, we include only retain samples that share that same entity. To do this, we cycle the forget samples of a forget target only over the retain samples of the same target (figure \ref{melu}). We will be left with general knowledge set to which we randomly assign a sample from the $D_f$.
\end{enumerate}
We adopt LoRA\cite{hu2022lora} for all our experiments. For finetuning, rank = 64, $\alpha$ = 128, batch size = 32 for 10 epochs. For unlearning, rank = 8, $\alpha$ = 16, batch size = 8\footnote{Batch size of 8 is maintained for every experiment by the aggregation of gradients
over 8 samples, even when the hardware limitations prohibit batch size of 8.} for 4 epochs. All the experiments were conducted on 2 x 40GB A100 GPUs. Full algorithmic details are in Appendix:\ref{appendix:methods}.

\subsection{Assessment}

Unlearning behavior is best assessed with the use of multiple metrics \cite{maini2024tofu}. We employ three distinct metrics and aggregate them to compute two scores: \textbf{Forget Efficacy} and \textbf{Model Utility}. In line with prior works \cite{maini2024tofu,yuan2025a,zhang2024negative}, we employ \textbf{ROUGE-L} (verbatim memorization with word-level match), \textbf{Conditional Probability} (ground truth likelihood) and \textbf{Cosine Similarity} (Semantic Similarity). To calculate \textit{Forget Efficacy}, we calculate \textit{1 - Arithmetic mean} of these metric on $D_f$ and for \textit{Model Utility} we calculate a harmonic mean of these metrics on $D_r$.

%% file: ECML_PKDD_2025_Author_Kit/results_discussion.tex
\section{Results and Discussion}

\textbf{Baselines} We computed Forget Efficacy (on forget set) and Model Utility (on test set) on the base model before unlearning. We use these results as baselines. After applying the unlearning algorithms, for a fair comparison, we use Forget Efficacy and Model Utility on Test set (MU-T), given that the $|D_r|$ changes based on the setting. The test set is balanced and will be used to understand the Model Utility. The base model has a low Forget Efficacy (FE = 0.30) and high model utility (ME-T = 0.73). A good unlearned model should have a higher FE (up to 1) and MU-T closer to the baseline $\pm{5}$. We also compute MMLU \cite{wang2024mmluprorobustchallengingmultitask} scores to understand the general Model utility, per target FE and MU-T for granular understanding and token diversity with \textbf{Distinct-N} \cite{li-etal-2016-diversity} to understand the diversity of the generated outputs.

\subsection{Evaluation of Unlearning Data Practices}
\subsubsection{Direct vs Indirect:} 
\begin{figure}[t] 
\includegraphics[width=\textwidth]{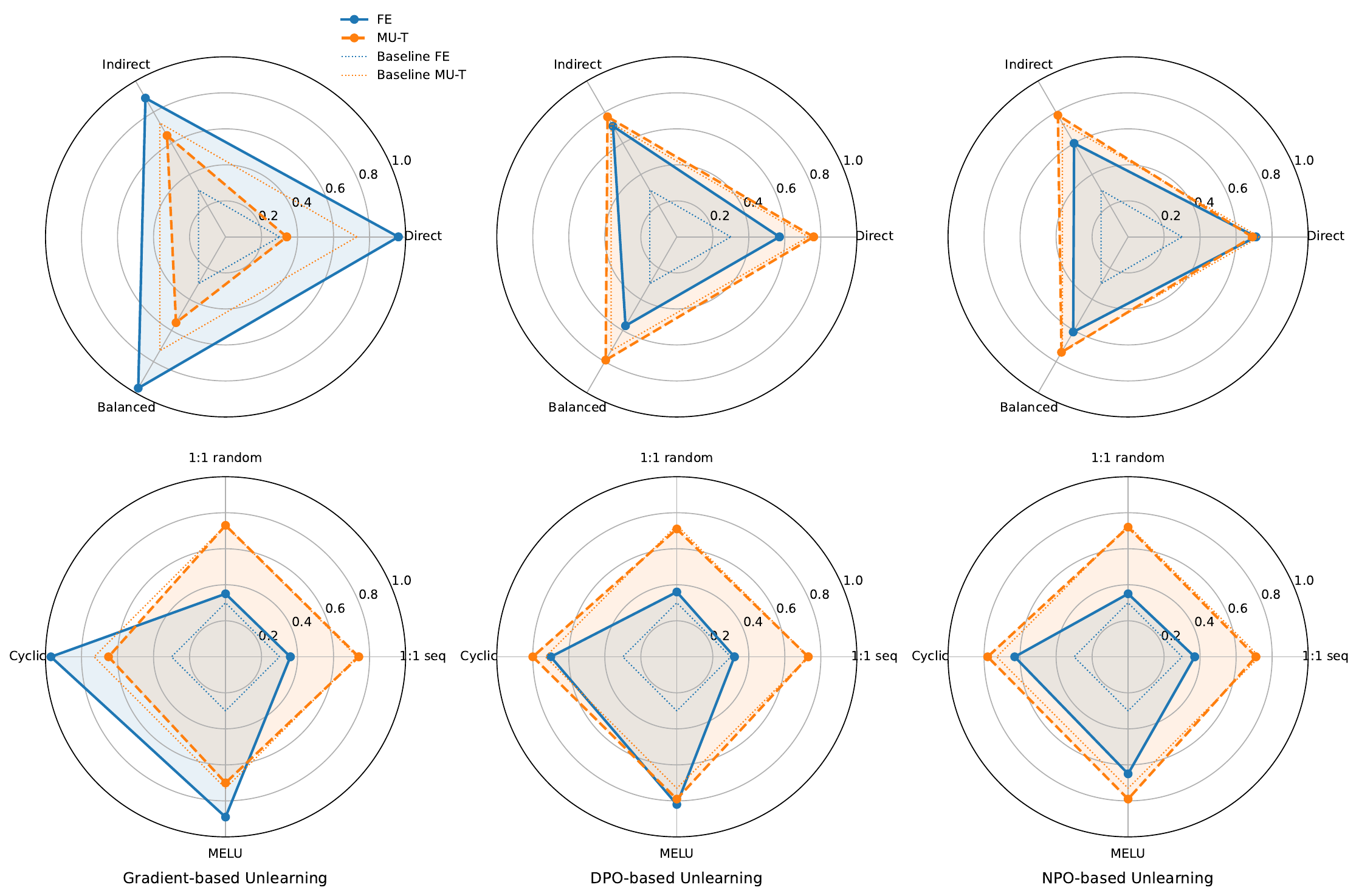}
\caption{FE and MU-T comparison results. Top row provides the data practices results and bottom row provides implementation practices results.} \label{results}
\end{figure}

Interestingly, both GD and NPO show (figure \ref{results} and table \ref{appendix:unlearning_results}) a drop in FE when moving from Direct $(\textit{N}_{d})$ to Indirect $(\textit{N}_{ind})$ neighbor sets, contrary to our expectations. Since, $(\textit{N}_{d})$ is smaller in size (\# 364 samples + \# 293 general knowledge) compared to $(\textit{N}_{ind})$ (\#1144 + \#293), we anticipated forget set would be revised more frequently during unlearning. However, $(\textit{N}_{d})$ neighbor with such a small neighbor set outperforms $(\textit{N}_{ind})$ in terms of FE. In contrast, DPO follows the expected trend, showing higher FE with the larger $(\textit{N}_{ind})$. However, \textbf{MU-T is consistently higher} for $(\textit{N}_{ind})$, indicates, larger and more diverse sets preserve general model performance. On the other hand, Balanced fails to achieve better FE and MU-T. 

\paragraph{}From the \textbf{Token Diversity} \ref{token_diversity} on the forget and test sets, we find that $(\textit{N}_{d})$ is lower than the $(\textit{N}_{ind})$ neighbors on the test set across all the unlearning algorithms. For GD, we see an exponential drop in forget set diversity in indirect and balanced. For DPO and NPO, we see an increment in token diversity from indirect to balanced. But both GD and DPO fail to maintain token diversity (even in implementation settings), this is because of GD's '\textit{Degeneration Behavior}' and DPOs '\textit{I don't know}' phrases. NPO exhibits more favorable behavior with high token diversity. This is likely due to its bounded objective, preventing model collapse. 

\paragraph{} \textbf{Per target FE and MU-T} show that GD, performs really well at forgetting with direct connections, but fails significantly at MU-T (10 targets are below 0.20 for MU-T). We find a similar situation with balanced, where the FE is higher and MU-T is lower. Although, indirect doesn't achieve same level of forgetting as direct, it always maintains $>0.85$ FE on all targets and maintains MU-T up to 0.65 (0.08 shy from baseline). With preference based methods, we find a gradual increment in MU-T from direct to indirect to balanced. A strange phenomenon was observed that some targets such as "\textit{Ted Kooser}" was harder to forget for both DPO and NPO (except direct case for NPO). With memorization scores (Appendix:Fig\ref{mem_Scores}), we find that these targets are highly memorized than others.

\paragraph{} \textbf{MMLU} scores on GD are inconsistent. Although they are not exponential drops or highs, but we find an increase in MMLU accuracy (Appendix:Fig\ref{appendix: mmlu_results}) on direct and balanced dataset experiments ($\approx +1.3\%$). This is an unusual behavior. Where as, preference based methods show a stable accuracy. Cyclic and MELU, provide stable MMLU scores across all the models, showing with proper implementation unlearning can be stable.

\subsection{Evaluation of Unlearning Sampling Practices}

Standard \textbf{1:1 sampling} (sequential and random) fails to produce meaningful forgetting (low FE) yet preserve MU-T across all the unlearning algorithms. Although, increase in number of epochs might improve the forgetting\footnote{To test this we conducted a run of DPO with 1:1 random sampling by continuously increasing the epochs. At epoch 100 we achieved 0.79 FE and 0.78 MU-T.}, we already achieve better stability (FE and MU-T) with cyclic and MELU with the same number of epochs. 

\paragraph{} \textbf{Stability with MELU} Both Cyclic and MELU perform significantly better than standard 1:1 sampling. They maintain stable performance across all the unlearning algorithms and maintain accuracy on MMLU and token diversity. MELU, in particular, outperforms cyclic under DPO, boosting FE by 12\% while maintaining MU-T. In NPO, MELU provides a small improvement from cyclic. But at the \textbf{per-target} performance, MELU holds a better FE and MU-T on all the targets for preference based methods. Under DPO, MELU increases the number of targets with FE > 0.9 (from 1 in cyclic to 3), while maintaining high MU-T \(\geq0.8\) for the majority. In case of Amy Clampitt, FE improves by $ \approx 20\%$. For GD, MELU setup achieves stable FE ($\approx 0.9$) across most entities and provides higher MU-T across targets. Even in NPO, where overall FE grows slowly, MELU maintains MU-T while achieving reasonable FE. In cases of harder targets such as "\textit{Ann Brashares}" and \textit{Ted Kooser}, both cyclic and MELU perform well and forget ($>0.50$) better than 1:1 sampling ($<0.10$). This improved stability of MELU can be attributed to a more consistent learning signal. In cyclic sampling, the model is subjected to high-variance gradients due to unrelated forget-retain pairs. MELU, by maintaining a relevancy between forget retain with lower variance per batch, could be leading to a stable performance. 

Overall MELU provides
\begin{enumerate}
    \item \textbf{High and Stable FE:} approaching or exceeding 0.85 for DPO and GD, and maintaining competitive scores in NPO.
    \item \textbf{Minimal degradation in MU-T:} consistently close to the baseline (0.73), even slightly exceeding it for some algorithms (e.g., NPO).
\end{enumerate}

%% file: ECML_PKDD_2025_Author_Kit/conclusion.tex
\section{Conclusion}

Our study demonstrates that the composition of the retain set is a critical, yet often under-looked factor in LLM Unlearning. Our findings show that relying solely on any single neighbor set is suboptimal and not the best practice. By including a diverse range of neighbors - we can improve the balance between forget efficacy and model utility. Furthermore, we show that the standard 1:1 sampling is an inefficient approach and when implementing unlearning, cyclic and Modular Entity-Level Unlearning (MELU) provides more stability. Albeit, we couldn't answer why these setups provides more stability, a research that could be looked into is if this caused due to the repetition of forget samples in the unlearning process (It is well-known that during the pretraining, memorization of a sample is correlated with its repetition in the corpus \cite{carlini2023quantifying}). 
Our work also proves some of the already proposed problems in the unlearning literature, such as Gradient based approach's instability \cite{wang2025rethinking,yao2024large}, and how some targets are harder to forget because of the frequency of their knowledge and memorization (Appendix:\ref{mem_Scores}) in pre-trained or downstream task data \cite{krishnan2025not}. We hope our work inspires researchers to look into methods to construct more diverse and realistic unlearning benchmarks and unlearning algorithm implementation techniques. 

For future works, We aim to conduct a rigorous comparative analysis against other model families, more unlearning algorithms and conduct a deeper evaluation, especially with consideration of sample memorization. Additionally, we aim to do a comparative analysis with Sequential Unlearning and also incorporating MELU into sequential unlearning.

\section{Limitations}


A key limitation of our study is the indirect neighbors and test sets generation with sister model. This limits the generalizability of our findings and requires expanding our experiments to other model families. Our MELU setup assumes that forget and retain samples are sufficiently distinct to be reliably separated. In real-world scenarios, however, such clear boundaries may not always exist, especially when entities share overlapping attributes or contexts, albeit one can enforce Knowledge graphs to define these clear connections\cite{qiu2025datainterconnectivityshapesllms}. Additionally, we do not deep dive into instability and stability issues such as GD and NPO's better forgetting on direct neighbors and not with indirect neighbors. In contrast DPO acts opposite, this can be further looked into especially through the lens of explainability approaches on pre and post unlearning. Same with MELU and cyclic settings stable performances, a rigorous work needs to be done towards addressing it. Another limitation is extension of general utility with HellaSwag \cite{zellers-etal-2019-hellaswag}, ARC \cite{chollet2025arcprize2024technical} etc. Our experiments cover only a few unlearning algorithms with batch unlearning. While we propose MELU, we lack a direct comparative analysis to sequential unlearning \cite{jang-etal-2023-knowledge} setups. Finally, because WPU \cite{liu-etal-2024-revisiting} already includes direct and general neighbors, we could not construct a full syntactic neighbor set, leaving it unexplored. 

\section*{Acknowledgments}
  This work has been partially supported by the EU EIC project \href{https://eic-emerge.eu}{EMERGE} (Grant No. 101070918).

%% file: ECML_PKDD_2025_Author_Kit/appendix.tex
\subsection{Prompts} \label{Prompts}

\subsubsection{Q\&A Generation}  \label{appendix:qa_generation}
\textit{This prompt was used to extract Q\&As from the wiki pages.}

\begin{tcblisting}{colback=white,
                  colframe=black,
                  title=LLaMA 3.3-70B,
                  breakable,
                  listing only}
#system_prompt\\
You are an expert teacher, who can create questions and answers from a given context. 
Given the user wikipedia page context about {# domain_person_name}, 

please provide as many questions and answers possible from it.

For each section, provide at least 2 questions and answers.

The question and answers should follow the Interrogative syntactic structure, 

The questions should be on their birth, family background, education, career, achievements and other relevant topics.

The output should be in JSON format with the following keys:

\{
    "name": name of the person,
    "question1": question1,
    "answer1": answer1,
    "section" : part of the wikipedia section,
    "difficulty" : difficulty of the question,
    "question2": question2,
    \dots
\}

Please be precise with the question and answer. Do not generate any other text.

#prompt\\
{# content}

\end{tcblisting}

\subsubsection{Indirect Connection generation}  \label{appendix:idrect_generation}
\textit{This prompt was used to generate six Indirect connections for a target.}

\begin{tcblisting}{colback=white,
                  colframe=black,
                  title=LLaMA 3.3-70B,
                  breakable,
                  listing only}

#prompt\\
For each name in the list, provide me 6 names that belong to the same domain as them (for example, if they are authors please provide authors similar as them). The output should be in a dictionary to make it into a dataframe.

['Benedetto Varchi', 'Wilhelm Wattenbach', 'Elsa Triolet', 'Theopompus', 'Heinrich Ritter', 'Adrienne Monnier', 'Ann Brashares', 'Hartmann von Aue', 'Jorge Semprún', 'Giovanni Battista Casti', 'Najaf Daryabandari', 'Heinz Erhardt', 'Rudolf Christoph Eucken', 'Paul Gerhardt', 'Moshe Greenberg', 'Amy Clampitt', 'Ted Kooser', 'Alfred Vogel', 'Siegfried Lenz', 'Philip Stanhope, 5th Earl Stanhope']

\end{tcblisting}


\subsection{Dataset} \label{dataset}
\begin{figure}[!htb]
\includegraphics[width=\textwidth]{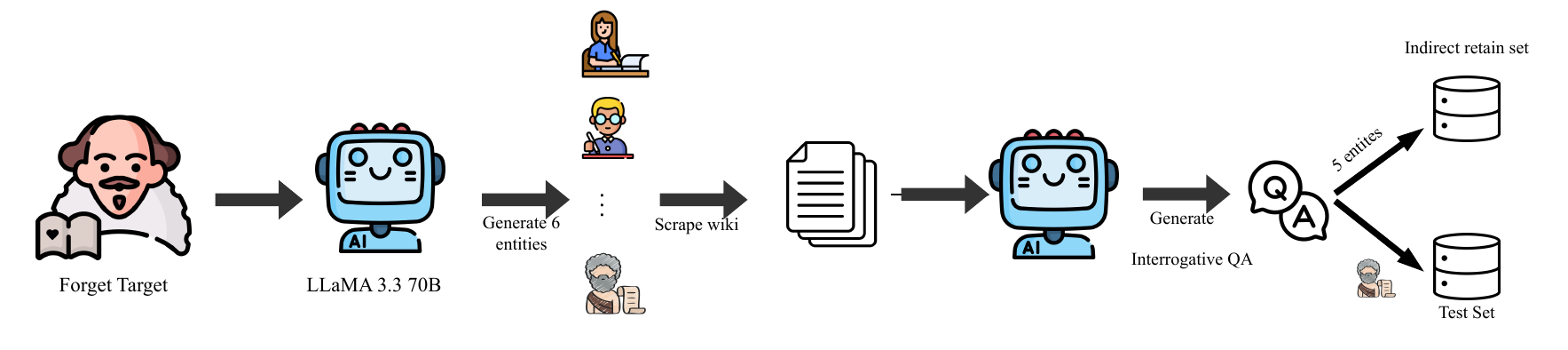}
\caption{Pipeline in creating Indirect connections for retain and test set} 
\label{appendix:indirect}
\end{figure}

A detailed pipeline in creating the indirect connections and its relevant test set samples are provided in the Figure \ref{appendix:indirect}. First, we use the prompt \ref{appendix:idrect_generation} to generate 6 entities for each target. Followed by we scrape their wiki pages and generate Q\&As with the LLaMA 3.3 70B model \ref{appendix:qa_generation} in an Interrogative syntactic structure manner to maintain the $(\textit{N}_s)$ neighbor dataset.
\vspace{-2em} 

\begin{figure}[!htb]
\includegraphics[width=\textwidth]{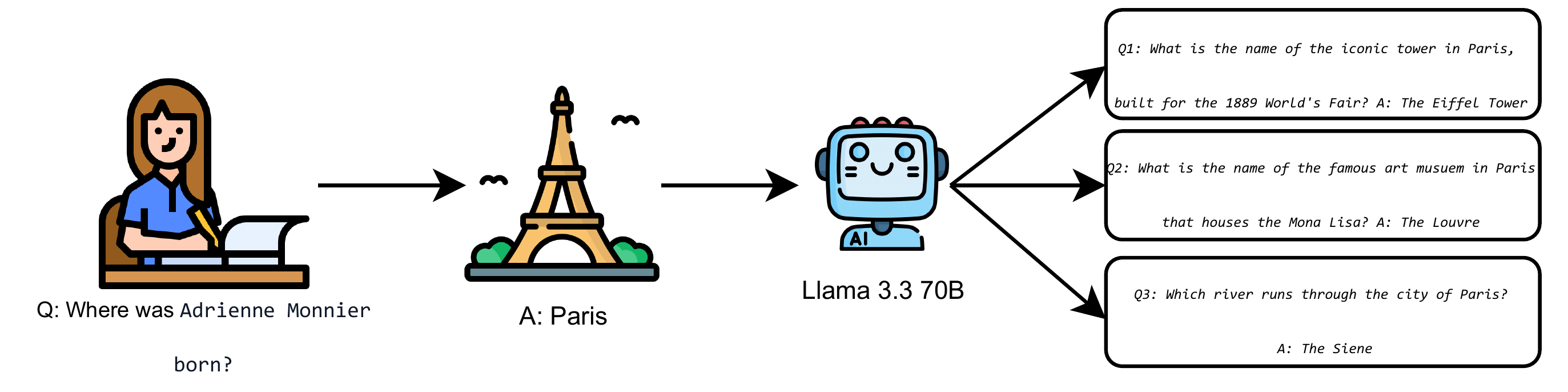}
\caption{Generation of Test set samples for the Direct Neighbors} 
\label{appendix:test_set}
\end{figure}

\begin{figure}[H]
  \centering
  \begin{minipage}[b]{0.48\textwidth}
    \includegraphics[width=\textwidth]{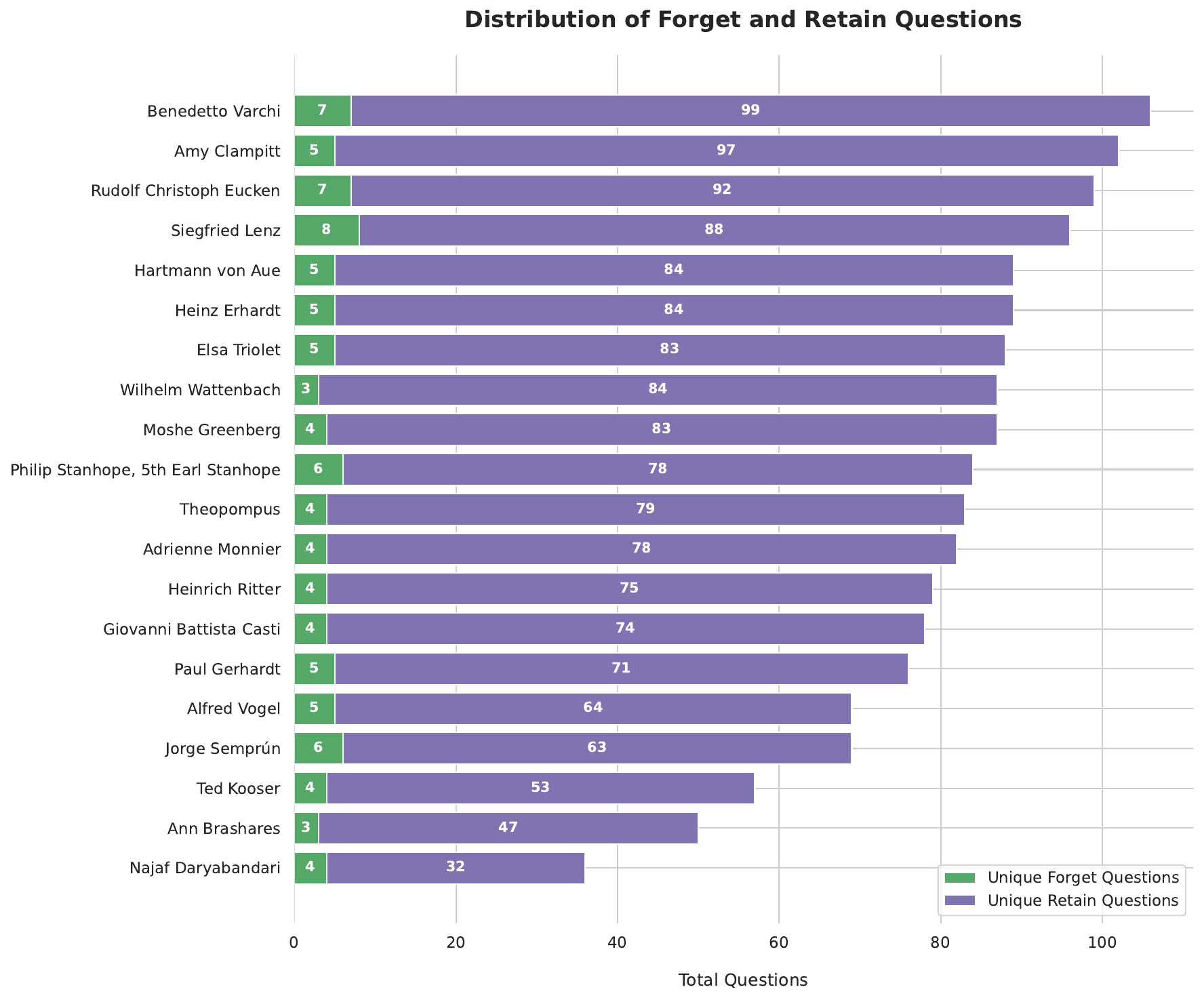}
    \caption{Composition of Forget‐Retain samples per target}
    \label{appendix:forget_retain_samples}
  \end{minipage}
  \hfill
  \begin{minipage}[b]{0.48\textwidth}
    \includegraphics[width=\textwidth]{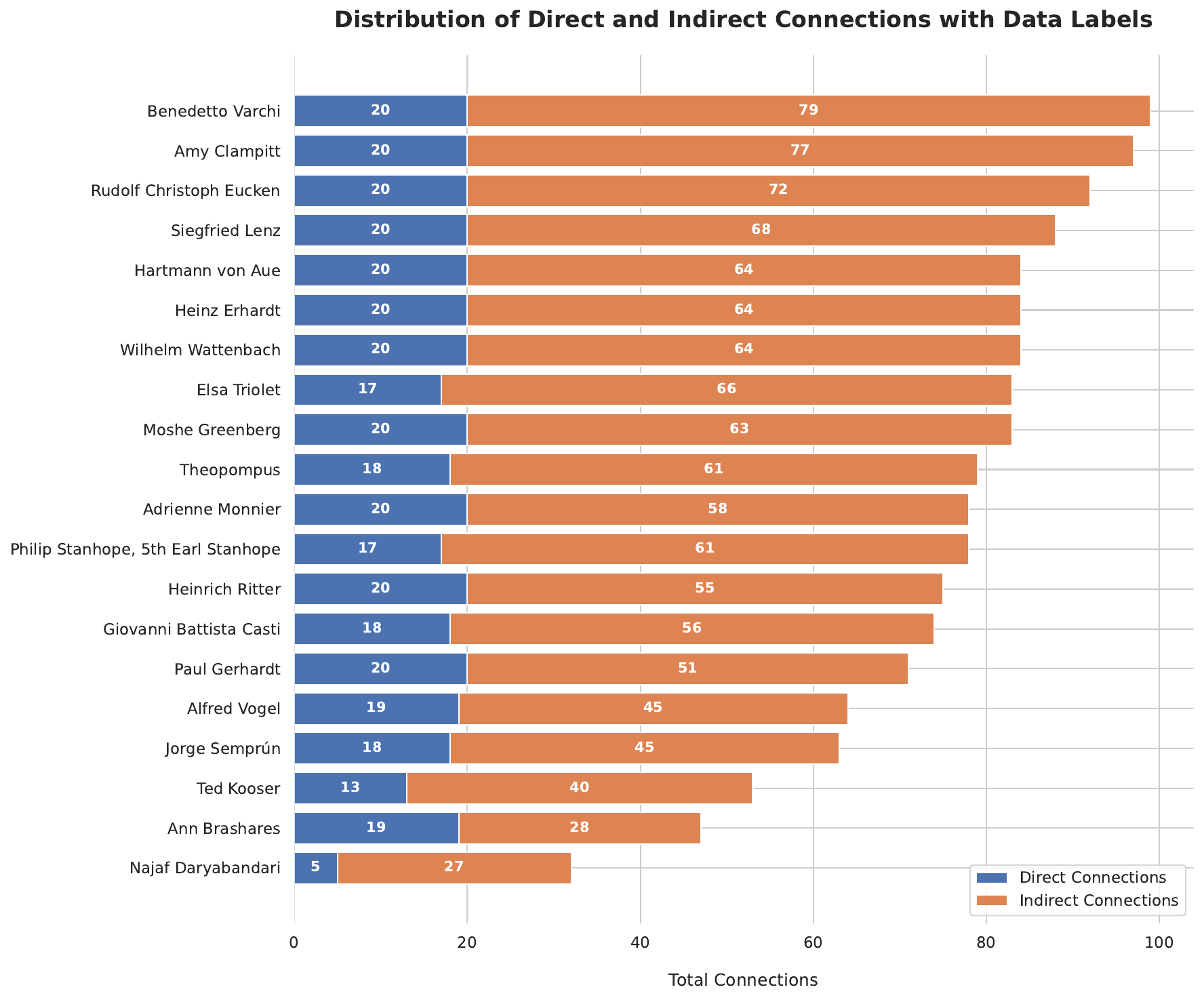}
    \caption{Composition of Direct and Indirect retain samples per target}
    \label{appendix:num_direct_indirect}
  \end{minipage}
\end{figure}

\subsection{Experimental setup} \label{appendix:methods}

\textbf{Model Finetuning}: Before the application of unlearning, we initially finetune the model on the all the datasets we have i.e., $D_f, D_r, D_t$. We use the questions as prompts and conduct a Supervised Fine-tuning on the datasets. Given $D_{ft} = D_f + D_r + D_t$, and its samples $(x,y)$, \textit{x} is question and \textit{y} is the answer. A pair $p_i = p(x_i,y_i) \in D_{ft}$ and $y_1,..y_T$ are the answer tokens, we calculate Negative-Log-Likelihood (NLL) loss for $p_i$

\begin{equation}
\mathcal{L}(y \mid x; \theta)
= \mathrm{NLL}(y \mid x; \theta)
= - \sum_{t=1}^{T} \log p\bigl(y_t \mid x, y_{<t}; \theta\bigr)
\label{eq:nll-definition}
\end{equation}

\subsubsection{Unlearning Methods:}
\paragraph{}
\textbf{Gradient Ascent (GA)}is the most straightforward unlearning technique proposed for the Un-targeted unlearning. It's main idea is to maximize the loss as opposed to the training objective of minimization by negating the loss. In our work, we do not implement this. Due to its nature of negation, the maximization becomes unbounded leading to catastrophic collapse. The predicted loss $l(y|x; \theta)$ on forget set can be written as:

\begin{equation}
\mathcal{L}_{GA}(D_f; \theta)
= -\,\mathcal{L}(y_f \mid x_f; \theta)
\label{eq:ga-loss}
\end{equation}

\paragraph{}
\textbf{Gradient Difference (GD)}proposed by \cite{liu2022continuallearningprivateunlearning} to mitigate the issues of Gradient ascent. It builds on the concept of Gradient Ascent, but not only aims to maximize the loss on forget set $D_f$, simultaneously minimizes the loss on the retain set $D_r$. This maintains the balance of forgetting and retaining. The loss function can be written as: 

\begin{equation}
\mathcal{L}_{GD}(\theta)
= \mathcal{- L}(D_f; \theta)
+ \mathcal{L}(D_r; \theta)
\label{eq:gd-loss}
\end{equation}

\paragraph{}
\textbf{Direct Preference Optimization (DPO)}is proposed by \cite{10.5555/3666122.3668460} and was first used by \cite{maini2024tofu}, treats unlearning as a preference optimization problem by applying the standard DPO loss. This technique uses \textit{Targeted Unlearning}, making a necessity of replacement responses like "I don't know". Alike the standard DPO approach, we use "I don't know" responses as positive samples and forget set as negative samples to guide the model's response. For the implementation of DPO, we convert the forget set to a preference dataset containing winning responses and refusal responses. Preference dataset $D_p = {(x_i, y_{i,win}, y_{i, lose})} , i \in |D_f|$, where $y_{i,win}$ are randomly chosen from a subset of "I don't know" phrases, and  $y_{i, lose}$ are the forget targets. 
The DPO loss can be calculated as:

\begin{equation}
\mathcal{L}_{DPO,\beta}(\theta)
= -
\,
\mathbb{E}_{D_p}\Biggl[
\log \sigma\Bigl(
\beta \log \frac{p(y_{\mathrm{win}}\mid x;\theta)}{p(y_{\mathrm{win}}\mid x;\theta_{\mathrm{ref}})}
\;-\;
\beta \log \frac{p(y_{\mathrm{lose}}\mid x;\theta)}{p(y_{\mathrm{lose}}\mid x;\theta_{\mathrm{ref}})}
\Bigr)
\Biggr]
\label{eq:dpo-beta-loss}
\end{equation}

where $\sigma$ is the sigmoid function and $\beta$ is the inverse temperature controlling the preference strength. We use $\beta = 0.1$ for all our experiments. Provided, we have a retain set, we utilize the code implementation provided by \cite{openunlearning2025}, can be calculated as follows:

\begin{equation}
  \mathcal{L}_{\mathrm{DPO+retain}}
  = \alpha\,\mathcal{L}_{\mathrm{DPO},\beta}(\theta)
  + \gamma\,\mathcal{L}(D_r;\theta)
  \label{eq:dpo-retain-loss}
\end{equation}

Where $\alpha$ and $\gamma$ are hyperparameter to control the strength of DPO loss and NLL. For our experiments, both $\alpha$ and $\gamma$ is always 1.

\paragraph{}
\textbf{Negative Preference Optimization (NPO) -} proposed by \cite{zhang2024negative}, is an inspiration of DPO, a variant that uses only the negative responses from the $D_f$, disregarding the $y_{win}$ making it an Un-targeted Unlearning. For the implementation, we ignore the "I don't know" responses and provide only $(x_f, y_f)$. Followed by, we calculate retain loss similarly as in DPO + retain with the same hyper-parameters values $(\beta, \alpha ,\gamma)$.

\begin{equation}
\mathcal{L}_{NPO,\beta}(\theta)
= -\frac{2}{\beta}
\,
\mathbb{E}_{D_p}\Biggl[
\log \sigma\Bigl(
\;-\;
\beta \log \frac{p(y_{\mathrm{lose}}\mid x;\theta)}{p(y_{\mathrm{lose}}\mid x;\theta_{\mathrm{ref}})}
\Bigr)
\Biggr]
\label{eq:NPO-loss}
\end{equation}

\begin{equation}
  \mathcal{L}_{\mathrm{NPO+retain}}
  = \alpha\,\mathcal{L}_{\mathrm{NPO},\beta}(\theta)
  + \gamma\,\mathcal{L}(D_r;\theta)
  \label{eq:dpo-retain-loss}
\end{equation}

\subsubsection{Evaluation Metrics:}
\setlength{\parskip}{0pt} 
\setlength{\parsep}{0pt}

\noindent\textbf{ROUGE (R)} quantifies the world-level overlap between the model's output and the ground-truth answer. We compute the ROUGE-L \cite{lin-2004-rouge} score between the generated response $g(x; \theta_*)$ and the ground-truth answer \textit{y}, written as $ROUGE-L(g(x; \theta_*), y)$. ROUGE-L provides the longest sequence overlap and the verbatim memory of the Unlearned Model $(M; \theta_*)$.
\medskip

\noindent\textbf{Cosine Similarity (CS)} measures the semantic similarity of the model's output against the ground-truth. We follow \cite{yuan2025a} setup, embed both with Sentence-BERT \cite{reimers-gurevych-2019-sentence}, calculate the cosine similarity and truncate the values less than 0.
\[
  \max\bigl(\cos\bigl(g(x;\theta_*),\,y\bigr),\,0\bigr)
\]


\noindent\textbf{Probability (P)} defines the average likelihood assigned to each token given a question and its ground truth answer i.e., $(x,y)$. Following  \cite{maini2024tofu}, we compute normalized conditional probability as
\[
  \mathcal{P}(y\mid x)
  = \tfrac{1}{T}\sum_{t=1}^T p\bigl(y_t\mid x\circ y_{<t};\,\theta_*\bigr)
\]

\newpage
\subsection{Results} 

{\setlength{\textfloatsep}{6pt plus 2pt minus 2pt}
 \setlength{\floatsep}{6pt plus 2pt minus 2pt}
 \setlength{\intextsep}{6pt plus 2pt minus 2pt}

 \captionsetup{font=small,skip=4pt}
 \captionsetup[table]{aboveskip=2pt,belowskip=2pt}
 \captionsetup[figure]{aboveskip=2pt,belowskip=2pt}
 
    \begin{table}[!ht]
    \centering
    \caption{Experimental results.}
    \label{appendix:unlearning_results}
    
    \scriptsize
    \setlength{\tabcolsep}{5pt}
    \renewcommand{\arraystretch}{0.90}
    
    \textbf{Dataset Practices}\\[3pt]
    \begin{tabular}{lccccc}
    \toprule
    Method & FE $\uparrow$ & MU‐T $\uparrow$ & PPL‐F $\downarrow$ & PPL‐T$\downarrow$ & MMLU \% \\
    \midrule
    Pre-Unlearning & 0.30 & 0.73 & 38.76 & 37105 & 12.42 \\
    \midrule
    \multicolumn{6}{l}{\textbf{Gradient-based (Un-Targeted)}} \\
    GA       & 0.44 & 0.67 & 657294.87 & 242062.34 & 12.47 \\
    Direct   & 0.96 & 0.34 & \(3.09 \times 10^{82}\) & \(3.28 \times 10^{80}\) & 13.60 \\
    Indirect & 0.89 & 0.65 & \(1.27 \times 10^{90}\) & \(1.79 \times 10^{82}\) & 8.40 \\
    Balanced & \textbf{0.97} & 0.55 & \(2.24 \times 10^{85}\) & \(1.79 \times 10^{82}\) & 13.29 \\
    \midrule
    \multicolumn{6}{l}{\textbf{DPO-based (Targeted)}} \\
    DPO      & 0.70 & 0.47 & 16643 & 3098 & 12.21 \\
    Direct   & 0.57 & 0.76 & \(1.82 \times 10^{4}\) & 158.72 & 12.27 \\
    Indirect & 0.71 & 0.77 & \(5.84 \times 10^{7}\) & 180.30 & 12.61 \\
    Balanced & 0.57 & 0.79 & \(1.71 \times 10^{5}\) & 142.58 & 12.35 \\
    \midrule
    \multicolumn{6}{l}{\textbf{NPO-based (Un-Targeted)}} \\
    NPO      & 0.30 & 0.73 & 38.76 & 37105 & 12.37 \\
    Direct   & \textbf{0.71} & 0.69 & \(4.68 \times 10^{22}\) & \(1.002 \times 10^{17}\) & 12.57 \\
    Indirect & 0.60 & 0.78 & \(3.17 \times 10^{18}\) & 126.47 & 12.37 \\
    Balanced & 0.61 & 0.74 & \(2.6 \times 10^{18}\) & 3153.19 & 12.63 \\
    \bottomrule
    \end{tabular}
    
    \vspace{0.5cm} 

    \textbf{Sampling Practices}\\[3pt]
    \begin{tabular}{lccccc}
    \toprule
    Method & FE $\uparrow$ & MU‐T $\uparrow$ & PPL‐F $\downarrow$ & PPL‐T$\downarrow$ & MMLU \% \\
    \midrule
    \multicolumn{6}{l}{\textbf{Gradient-based (Un-Targeted)}} \\
    1:1 seq     & 0.36 & \textbf{0.74} & \(3.7 \times 10^{4}\) & \(6.11 \times 10^{4}\) & 12.46 \\
    1:1 random  & 0.35 & 0.73 & 22521 & 32751 & 12.31 \\
    Cyclic      & \textbf{0.97} & 0.65 & \(1.80 \times 10^{86}\) & 3.08 & 13.26 \\
    MELU        & 0.89 & 0.70 & \(3.18 \times 10^{90}\) & 15.98 & 13.42 \\
    \midrule
    \multicolumn{6}{l}{\textbf{DPO-based (Targeted)}} \\
    1:1 seq     & 0.32 & 0.73 & 124.55 & \(3.0 \times 10^{3}\) & 12.33 \\
    1:1 random  & 0.36 & 0.71 & 65.90 & 2539.52 & 12.28 \\
    Cyclic      & 0.70 & \textbf{0.80} & \(2.57 \times 10^{7}\) & 118.54 & 12.36 \\
    MELU        & \textbf{0.82} & 0.79 & \(5.8 \times 10^{14}\) & 87.71 & 12.38 \\
    \midrule
    \multicolumn{6}{l}{\textbf{NPO-based (Un-Targeted)}} \\
    1:1 seq     & 0.37 & 0.71 & 1655.97 & \(5.33 \times 10^{5}\) & 12.43 \\
    1:1 random  & 0.35 & 0.72 & 14545.83 & 78544 & 12.36 \\
    Cyclic      & 0.63 & 0.78 & \(1.60 \times 10^{18}\) & 35.15 & 12.24 \\
    MELU        & 0.65 & \textbf{0.79} & \(7.86 \times 10^{21}\) & 54.03 & 12.41 \\
    \bottomrule
    \end{tabular}
    
    \end{table}

    \vspace{1cm}
    \noindent\begin{minipage}{\linewidth}
    \centering
    \includegraphics[width=0.85\linewidth]{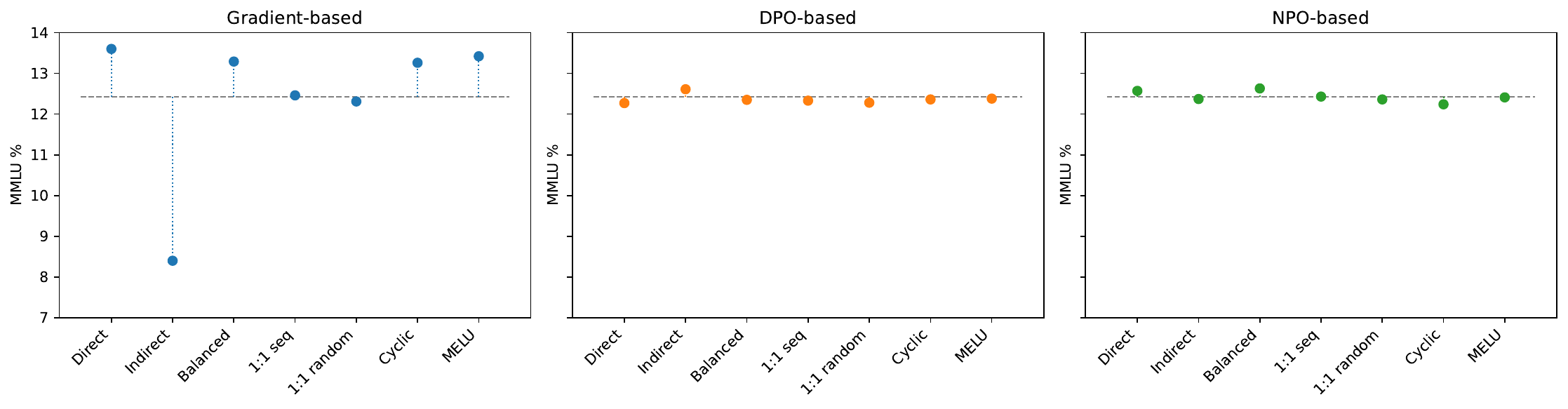}
    \captionof{figure}{General Model Utility (MMLU) across all experiments. Baseline accuracy is 12.42\%.}
    \label{appendix: mmlu_results}
    \end{minipage}
}

\begin{figure}[H]  
  \centering

  \begin{subfigure}{0.95\linewidth}
    \centering
    \includegraphics[width=\linewidth,height=0.26\textheight,keepaspectratio]
      {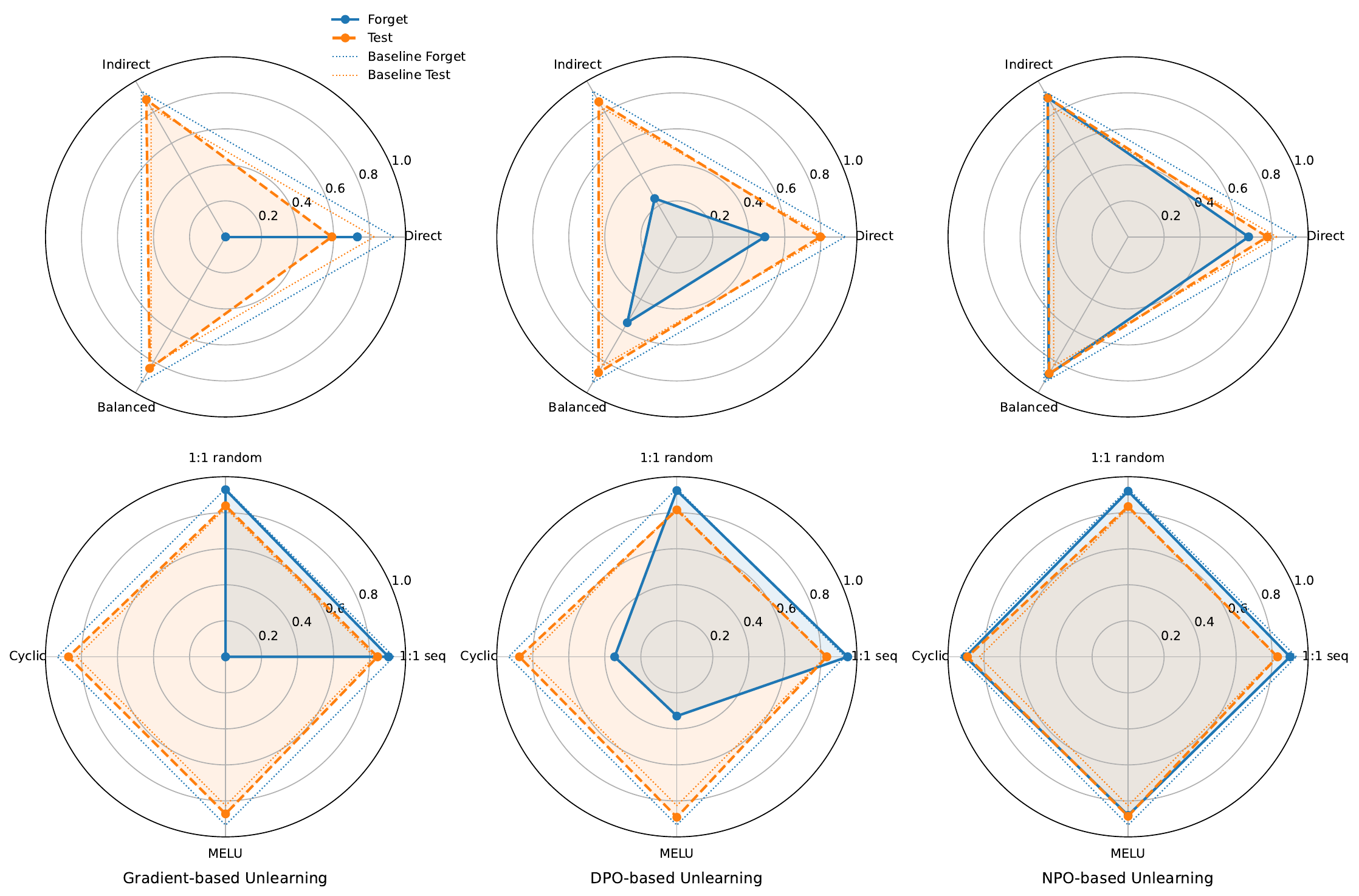}
    \caption{Token diversity of the unlearned models. Top row: data practices; bottom row: implementation practices.}
    \label{token_diversity}
  \end{subfigure}

  \vspace{0.6em}
  {\bfseries Heat maps of per-entity FE and MU-T}\par
  \vspace{0.3em}

  \begin{subfigure}{0.9\linewidth}
    \centering
    \includegraphics[width=\linewidth,height=0.26\textheight,keepaspectratio]
      {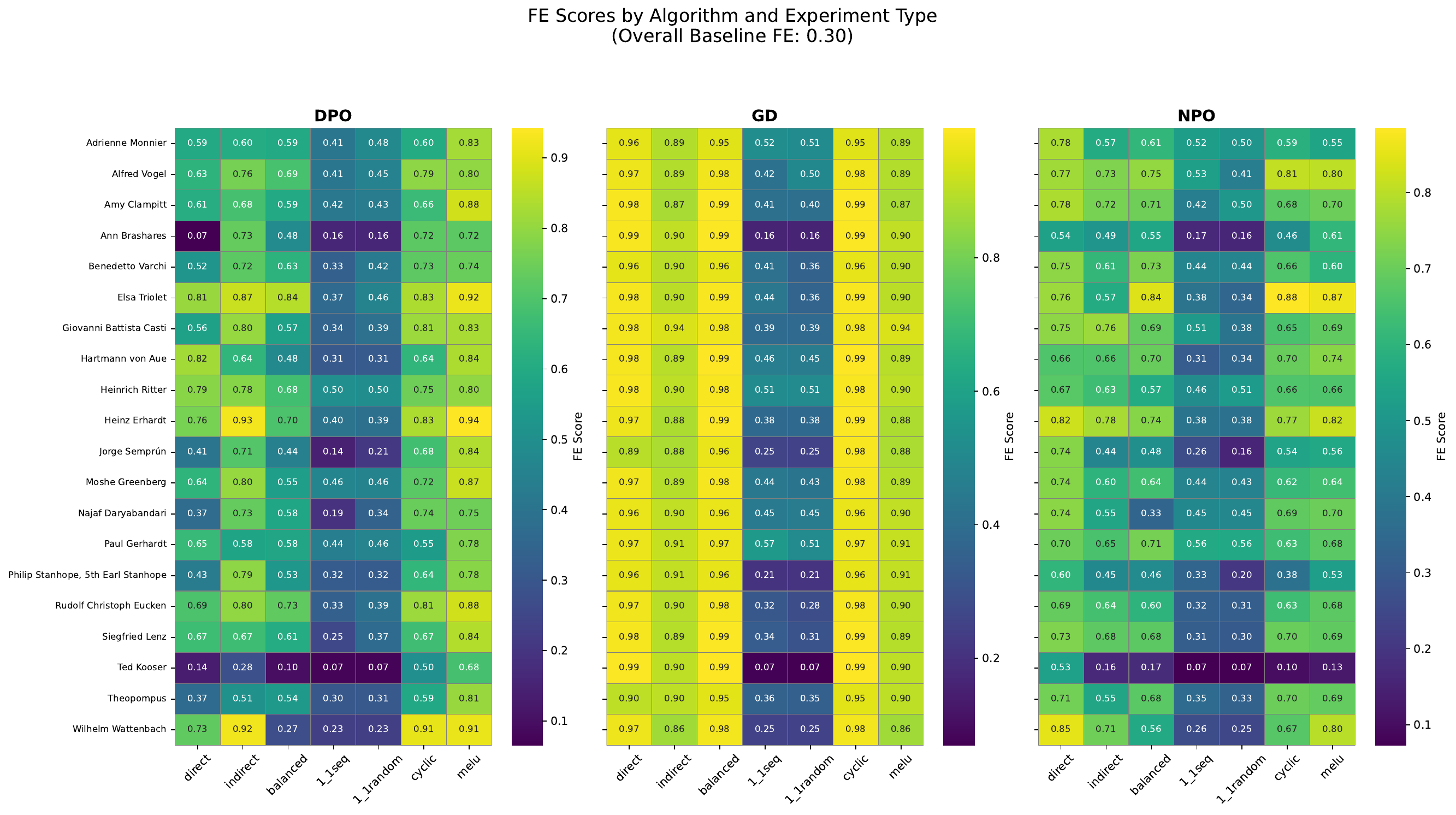}
    \caption{Forget Efficacy (FE).}
    \label{per_title_fe}
  \end{subfigure}

  \vspace{0.6em}

  \begin{subfigure}{0.9\linewidth}
    \centering
    \includegraphics[width=\linewidth,height=0.26\textheight,keepaspectratio]
      {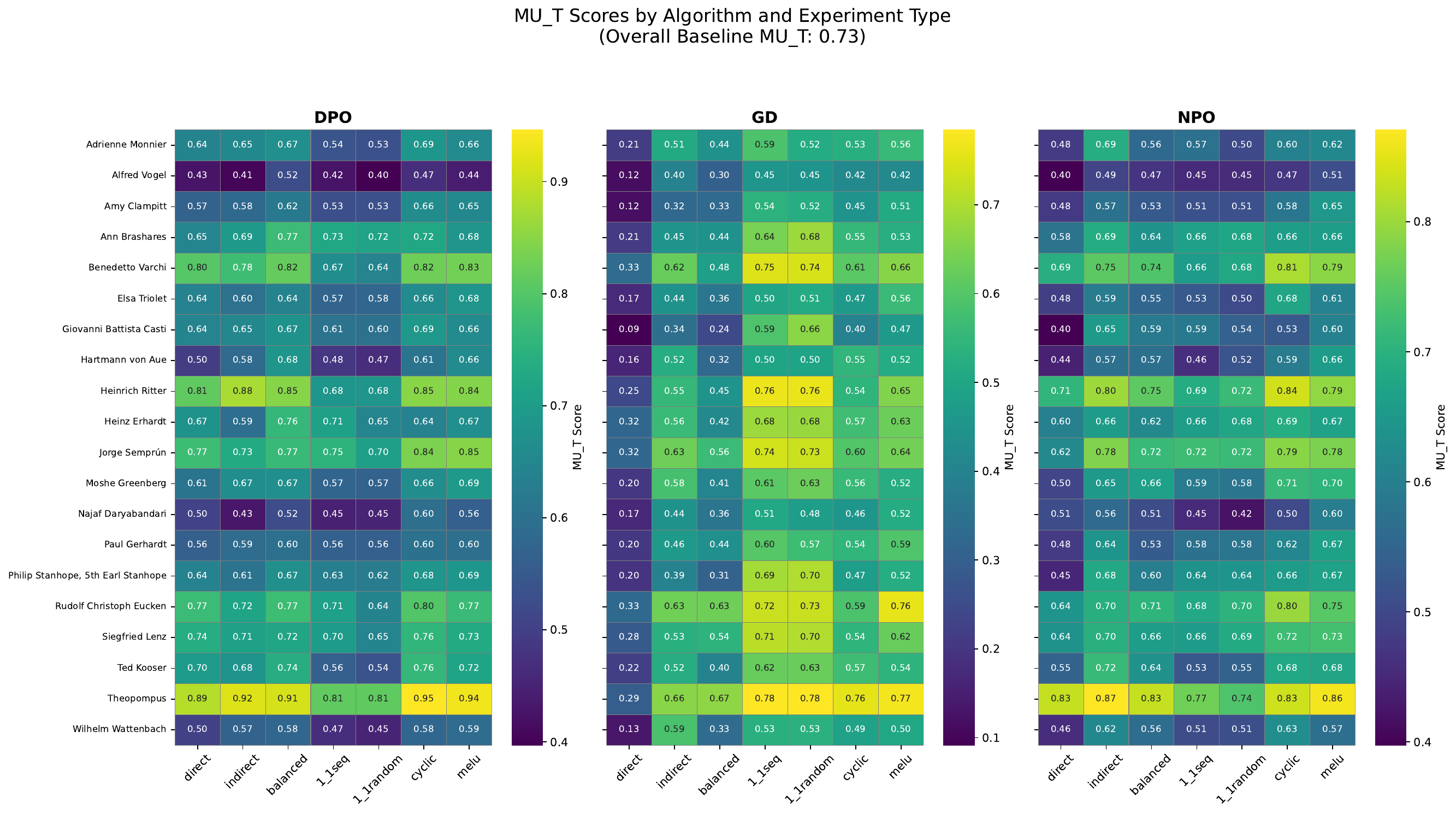}
    \caption{Model Utility Test (MU-T).}
    \label{per_title_mu_t}
  \end{subfigure}

  \caption{Token diversity and per-entity metrics.}
  \label{fig:combined}
\end{figure}

\FloatBarrier

\begin{figure}[H] 
\centering
\includegraphics[width=\linewidth, height=0.42\textheight, keepaspectratio]{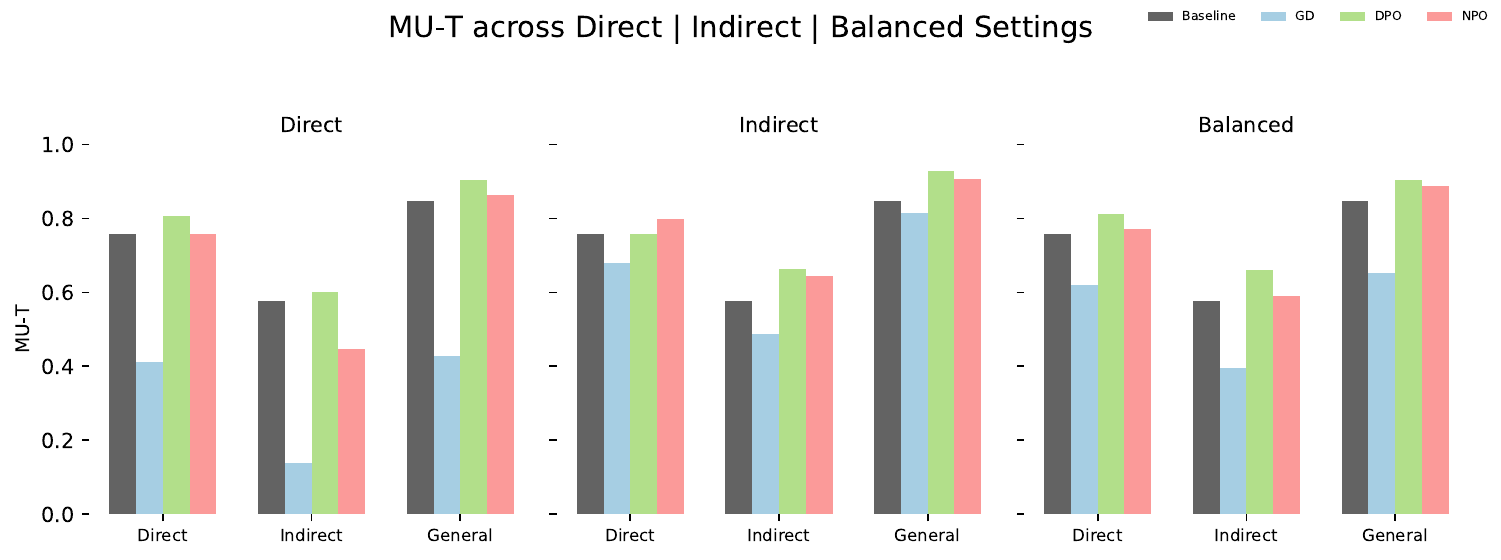}
\caption{MU-T across various data practices. We find Gradient Based method consistently performs bad in all the Data settings, where as NPO and DPO holds the MU-T well.}
\label{mut-datasets}
\end{figure}
\FloatBarrier

\subsection{Memorization}

We calculated each entities memorization with Exact Memorization (EM) score following \cite{tirumala2022memorization}, which is often used in Unlearning research to define/compute the success of forgetting in LLMs. Given we have multiple samples for each entity, we compute their Average EM score. We find that few samples such as \textit{Ted Kooser, Philip Stanhope, Ann Brashes etc} are highly memorized and were harder to forget in our experiments.

\begin{figure}[H] 
\centering
\includegraphics[width=\linewidth, height=0.42\textheight, keepaspectratio]{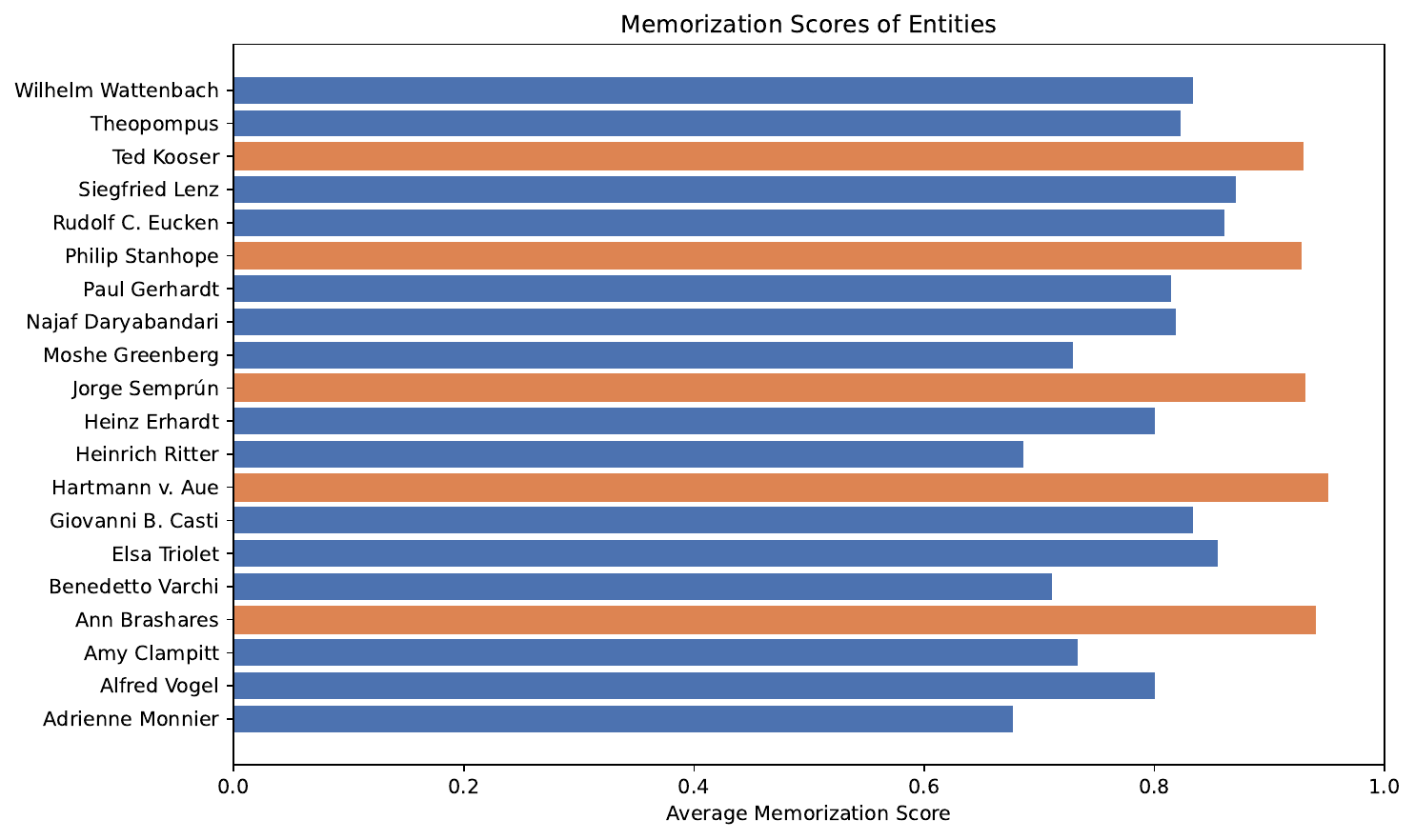}
\caption{Memorization Scores of Each Entity}
\label{mem_Scores}
\end{figure}
\FloatBarrier